\documentclass[letterpaper, 10 pt, conference]{ieeeconf}  %

\IEEEoverridecommandlockouts                              
\usepackage[utf8]{inputenc}
\usepackage[T1]{fontenc}
\usepackage{cite}
\usepackage[table]{xcolor}
\definecolor{rblue}{rgb}{0,0.5,1}
\definecolor{awesome}{rgb}{1.0, 0.13, 0.32}
\definecolor{hollywoodcerise}{rgb}{0.96, 0.0, 0.63}
\definecolor{lasallegreen}{rgb}{0.03, 0.47, 0.19}
\definecolor{hanpurple}{rgb}{0.32, 0.09, 0.98}
\definecolor{green(pigment)}{rgb}{0.0, 0.65, 0.31}

\makeatletter
\let\NAT@parse\undefined
\makeatother
\usepackage[pagebackref=false, breaklinks=true, colorlinks, bookmarks=false]{
        hyperref
}
\hypersetup{
        colorlinks=true,
        linkcolor={red},
        citecolor={hanpurple},
        urlcolor={magenta}
}

\usepackage{caption}
\usepackage{graphicx}
\usepackage{amsmath}
\usepackage{amssymb}
\usepackage{booktabs}

\title{\LARGE \bf
PanoFuse: Panorama-Enhanced Vision-Language-Action Learning with Decoupled Semantic-Geometric Routing}

\author{Peng Xu$^{1}$, Haoran Lin$^{1}$, Wanjun Jia$^{1}$, Kai Luo$^{1}$, Wenrui Chen$^{1,2}$, Zhiyong Li$^{1,2}$, and Kailun Yang$^{1,2,\dag}$
\thanks{This work was supported in part by the National Natural Science Foundation of China (Grant No. 62473139 and No. 62388101), in part by the Hunan Provincial Research and Development Project (Grant No. 2025QK3019), and in part by the State Key Laboratory of Autonomous Intelligent Unmanned Systems (the opening project number ZZKF2025-2-10).}
\thanks{$^{1}$The authors are with the School of Artificial Intelligence and Robotics, Hunan University, China (email: kailun.yang@hnu.edu.cn).}%
\thanks{$^{2}$The authors are also with the National Engineering Research Center of Robot Visual Perception and Control Technology, Hunan University, China.}%
\thanks{$^{\dag}$Corresponding author: Kailun Yang.}
}

\let\oldtwocolumn\twocolumn
\renewcommand\twocolumn[1][]{%
    \oldtwocolumn[{#1}{
    \begin{center}
    \vskip -3ex
        \centering
        \includegraphics[width=0.99\textwidth]{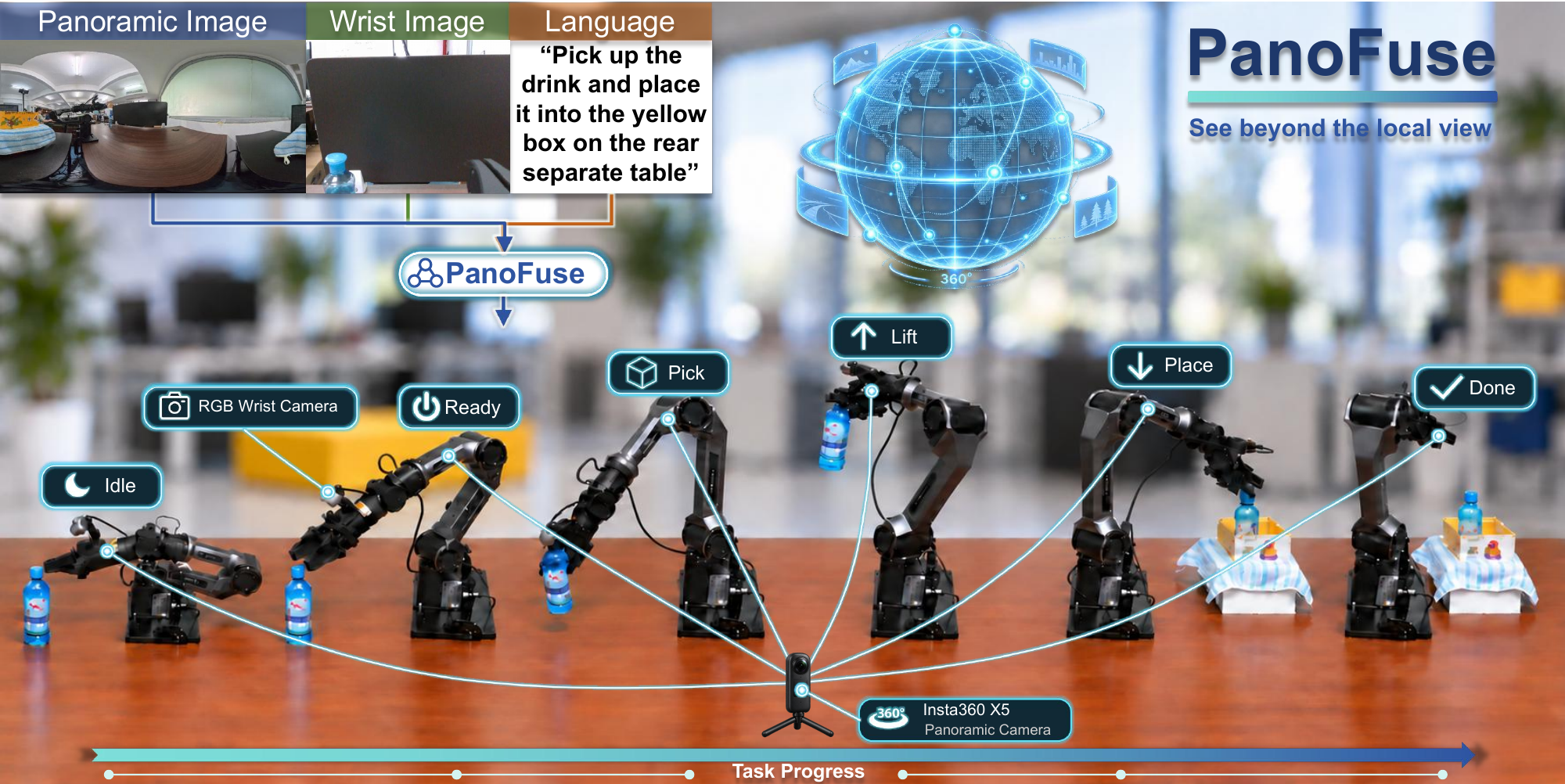}
        \vskip -1ex
        \captionof{figure}{{Overview of PanoFuse for real-world robotic manipulation. PanoFuse complements local wrist-camera observations with a $360^\circ$ panoramic view to provide global scene context. The illustration shows a representative long-horizon manipulation sequence, where panoramic perception supports continuous action generation from target acquisition and grasping to transfer and place.
        }}
        \label{fig:overview}
    \end{center}
    }]
}

\begin{document}

\maketitle
\thispagestyle{empty}
\pagestyle{empty}

\begin{abstract}
Vision-Language-Action (VLA) policies have shown promising performance in language-conditioned robotic manipulation. However, most existing VLA systems rely on conventional perspective cameras with limited fields of view, often missing global scene context and leading to unreliable manipulation under visual occlusions, distractors, and unseen environments. In this work, we propose \textbf{PanoFuse}, a panorama-enhanced VLA framework that complements local manipulation observations with global panoramic perception. PanoFuse introduces a dedicated panoramic branch that leverages a pretrained panoramic foundation model to extract complementary semantic and geometric representations from omnidirectional observations. Rather than directly mixing these heterogeneous features, we introduce Decoupled Semantic-Geometric Routing (DSGR), which maintains semantic and geometric representations as separate context streams and selectively routes both to downstream state and action representations through structured block-wise attention. This design provides the action expert with global spatial context while preserving task-relevant semantic information from the pretrained VLA backbone. We further develop a synchronized data collection pipeline and construct a new real-world manipulation dataset containing panoramic RGB observations, wrist-view images, language instructions, robot states, and actions. Across seven evaluation settings, PanoFuse achieves an average success rate of $52.9\%$, outperforming the evaluated baselines and achieving consistent gains under novel-object, unseen-background, and distractor-rich settings. Code and data will be released publicly at \url{https://xux-hnu.github.io/PanoFuse}.
\end{abstract}

\section{Introduction}

Vision-Language-Action (VLA) models have recently shown strong potential for general-purpose robotic manipulation by combining large-scale visual-language representations with action generation~\cite{kim2024openvla,black2025pi_0,octo_2023,wen2025dexvla,zhao2025cot}. 
Leveraging pretrained Vision-Language Models (VLMs), these methods inherit strong semantic understanding and language grounding capabilities, enabling a single policy to generalize across diverse manipulation tasks and robot embodiments.

Despite this progress, robust manipulation in real-world environments remains fundamentally limited by what the robot can observe. 
Most existing VLA systems rely on one or several perspective cameras with relatively narrow fields of view~\cite{kim2024openvla,black2025pi_0,octo_2023,wen2025dexvla,zhao2025cot}. As a result, the policy only observes a local portion of the workspace at each timestep. Task-relevant objects may lie outside the camera frustum, become occluded by the robot arm or surrounding objects, or disappear from view as the robot moves\cite{qu2026omnidp}. These challenges are particularly pronounced in cluttered and unstructured environments, where reliable manipulation requires maintaining task-relevant spatial context beyond the instantaneous local view\cite{yang2026learning}.

This limitation reveals a mismatch between the reasoning capability of modern VLA models and their sensing range: 
\emph{the policy may reason globally in semantics, while perceiving only locally in space}. In everyday manipulation, humans naturally exploit a broad view of the surrounding workspace to maintain awareness of targets, obstacles, and free space, even under partial occlusion. Similarly, wide-field and panoramic perception can provide robots with more complete scene coverage without requiring repeated camera or body reorientation. Such global sensing is especially valuable in cluttered manipulation scenarios, where partial occlusions, distractors, unseen backgrounds, or limited camera coverage can hinder reliable perception.

Several recent works~\cite{yuan2025depthvla,lan2025bfa,yang2026multi_view} attempt to alleviate partial observability through depth-enhanced perception or multi-view sensing. 
However, these approaches often introduce additional system complexity, such as multi-camera calibration, extra exploratory motions, or dependence on reconstructed or hallucinated scene content. More importantly, when the sensing setup remains dominated by conventional perspective cameras, the observable workspace is still fundamentally restricted by the camera field of view.

Panoramic perception offers a direct way to address this limitation at the sensing level by providing continuous global observation of the workspace. However, directly applying pretrained vision-language encoders to panoramic observations remains challenging, as equirectangular images exhibit substantial geometric distortion and contain large amounts of task-irrelevant visual content~\cite{Ling_2023_CVPR}. A manipulation policy therefore needs to extract task-relevant global structure from the panorama and integrate it with the pretrained semantic features of the VLA backbone in an efficient manner.

To address these challenges, we propose PanoFuse, a panorama-enhanced VLA framework for robust robotic manipulation. As illustrated in
Fig.~\ref{fig:overview}, PanoFuse complements local wrist-camera observations with a $360^\circ$ panoramic view of the surrounding workspace, enabling the policy to maintain global scene awareness throughout long-horizon manipulation. PanoFuse extracts complementary semantic and geometric representations from the panoramic observation and maintains them as separate context streams. We further introduce Decoupled Semantic-Geometric Routing (DSGR) to selectively route these heterogeneous representations toward downstream state and action representations through structured block-wise attention. This design enables the policy to exploit global spatial context while preserving the semantic priors and action-generation capability of the
pretrained VLA backbone.

We evaluate PanoFuse on seven evaluation settings covering cross-view manipulation, sequential manipulation, and generalization under novel objects, unseen backgrounds, and visual distractors. 
The collected dataset contains $200$ successful teleoperated trajectories and over $100K$ synchronized frames. 
In closed-loop real-robot evaluations, PanoFuse achieves an average success rate of $52.9\%$, substantially outperforming the best evaluated baseline at $30.0\%$.
These results demonstrate the effectiveness of panoramic context for improving robustness and generalization in real-world manipulation.

Our main contributions are summarized as follows:
\begin{itemize}
    \item We propose PanoFuse, a panorama-enhanced VLA framework that provides global scene perception and alleviates partial observability caused by limited camera fields of view.

    \item We introduce DSGR that integrates pretrained VLM semantics with global panoramic geometry through structured block-wise attention.

    \item We build a panorama-based robotic manipulation dataset and evaluate PanoFuse on seven evaluation settings. PanoFuse achieves an average success rate of $52.9\%$, outperforming the evaluated baselines and demonstrating consistent improvements across standard and generalization settings.
    
\end{itemize}

\begin{figure*}[!t]
    \centering
    \includegraphics[width=\textwidth]{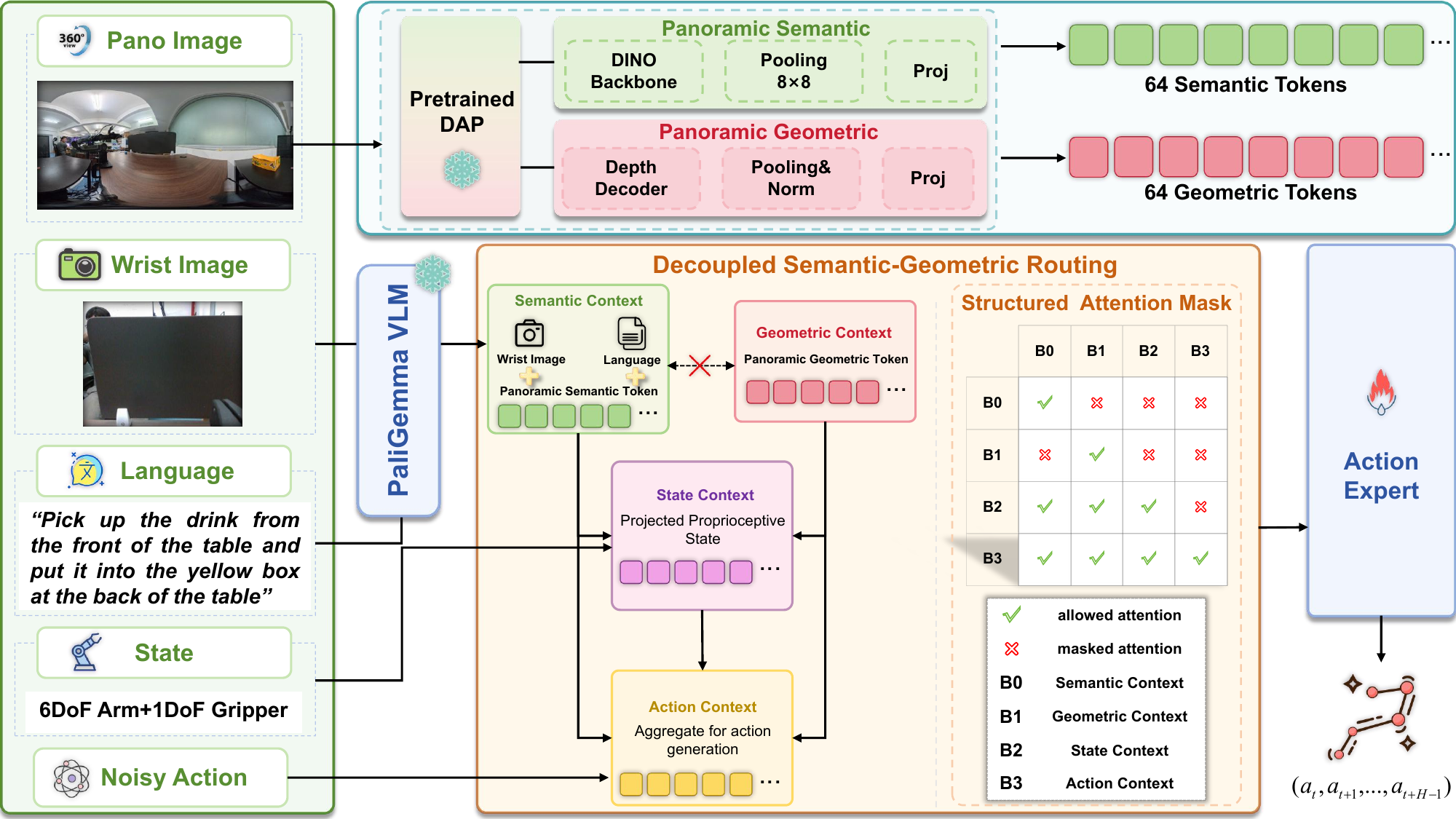}
    \vskip-1ex
    \caption{\textbf{Pipeline of the proposed PanoFuse framework.}
    (1) \textbf{Panoramic Representation:} 
    A frozen Depth Any Panoramas (DAP) model~\cite{lin2026depth} extracts complementary semantic and geometric tokens from panoramic observations.
    (2) \textbf {Decoupled Semantic-Geometric Routing:} Semantic and geometric contexts are separately contextualized and selectively routed to downstream state and action representations through structured attention.
    (3) \textbf{Action Generation:} The action expert conditions on the routed multimodal context to generate the final action chunk.}
    \label{fig:pipeline}
    \vskip-3ex
\end{figure*}

\section{Related Work}
\subsection{Vision-Language-Action Systems}
Vision-Language-Action (VLA) models aim to learn generalizable robotic policies that directly generate executable actions from visual observations and language instructions. 
Early approaches~\cite{kim2024openvla,pmlr-v229-zitkovich23a} commonly formulate action prediction as autoregressive generation over discrete action tokens, while recent methods~\cite{black2025pi_0,jiang2026asyncvlaasynchronousflowmatching} increasingly adopt generative paradigms such as flow matching to model continuous, multimodal action distributions and produce smooth action sequences. 
Beyond action generation, recent VLA research~\cite{yuan2025depthvla,lan2025bfa,yang2026multi_view} has explored richer perceptual representations to improve robustness and generalization in complex manipulation environments. A growing body of work has investigated the limitations of visual perception in VLA systems. Some methods~\cite{an2026uncovering_blind_spots,pan2026vla_corrector,pu2026tgm_vla} explicitly address visual blind spots and incomplete observations, while others~\cite{li2026libero_occ,yu2025forcevla,syed2026intercepting,liu2026rocap} focus on robustness to occlusion through complementary sensing, predictive modeling, or recovery mechanisms. 
Stereo-based approaches~\cite{wu2026eatr,deng2025stereovla} further incorporate binocular observations to enhance geometric and spatial reasoning. More recently, multi-view VLA methods~\cite{yang2026multi_view,zhang2026_4d,zhao2026_gwm_vla,xiao2026learning_action_manifold,bian2025_vla_lpaf,li2025_uni_sight,zhang2026_oc_vla,ling2026guide} aggregate observations from multiple cameras or viewpoints to provide broader scene coverage and richer spatial context. However, most existing VLAs still rely on local pinhole or discrete multi-view observations, which provide incomplete scene coverage and may miss task-relevant context.

\subsection{Panorama-Based Perception in Robotics}
Panoramic cameras provide nearly complete spherical coverage ($360^\circ \times 180^\circ$), enabling global observation of the surrounding environment from a single viewpoint~\cite{gao2022review}.
Compared with conventional pinhole cameras with limited fields of view, panoramic vision provides substantially broader scene context and reduces perceptual blind spots. Benefiting from this property, panoramic imaging has been widely explored for object detection~\cite{wang2009object}, semantic segmentation~\cite{yang2019pass}, and 3D geometric understanding~\cite{zhang2021deeppanocontext}. 
In robotics, panoramic perception has also been applied to navigation~\cite{jin2026panonav,arif2024panoramic_visual,wang2026monodream,li2025between,cai2026navigation_wayfinding} and visual tracking~\cite{bacchin2022people}.

Despite these advances, panoramic perception remains relatively under-explored for vision-language-action policy learning. Recent studies have begun to incorporate omnidirectional perception into robotic 
manipulation. OmniDP~\cite{qu2026omnidp} employs panoramic LiDAR observations within an end-to-end visuomotor policy, leveraging omnidirectional 3D geometry to enable manipulation beyond the field of view (FoV) over large workspaces. 
PanoVLA~\cite{yang2026learning} incorporates panoramic observations into a VLA architecture through dedicated panorama encoding and fusion modules. Our work investigates how semantic and geometric representations extracted from panoramic observations can be organized within a pretrained VLA policy. Specifically, PanoFuse maintains these representations as separate context streams and uses structured attention to expose both streams to state and action tokens.

\section{Methodology}
We first formulate the language-conditioned robotic manipulation problem with local and global visual observations (Sec.~\ref{sec:Problem}). As illustrated in Fig.~\ref{fig:pipeline}, PanoFuse extracts complementary panoramic semantic and geometric representations and integrates them with wrist-view, language, state, and action contexts through Decoupled Semantic-Geometric Routing (DSGR) to condition the action expert. We then detail the panoramic representation, DSGR mechanism, and training objective in Sec.~\ref{sec:PanoFuse}.

\subsection{Problem Formulation}
\label{sec:Problem}
We consider language-conditioned robotic manipulation with both local and global visual observations. To complement the limited field of view of conventional local cameras, we supplement robot observation with a panoramic RGB view that captures the surrounding workspace and provides a global context of the scene. Specifically, at time step $t$, the observation is defined as
\begin{equation}
    o_t = \{I_t^{wrist}, I_t^{pano}, s_t\}
    \label{eq:observation}
\end{equation}
where $I_t^{wrist}$ denotes the local wrist-view RGB observation, $I_t^{pano}$ denotes the panoramic RGB observation covering the surrounding workspace, and $s_t$ represents the robot's proprioceptive state.
Given the observation $o_t$ and a language instruction $l$, the policy $\pi_\theta$ predicts a horizon-H continuous action chunk,
\begin{equation}
    a^{t:t+H-1} = \pi_{\theta}(o_t, l),
    \label{eq:continuous_action_chunk}
\end{equation}
where each action $a_t = [q_t^{cmd} ,g_t^{cmd}] \in R^7$ consists of the commanded positions $q_t^{cmd} \in R^6$ of the six arm joints and a gripper command $g_t^{cmd}$.

\subsection{PanoFuse}
\label{sec:PanoFuse}
\noindent\textbf{Model Architecture.} 
PanoFuse builds upon the pretrained $\pi_0$ policy and augments it with a dedicated panoramic branch while preserving its pretrained vision-language pathway. Rather than directly fusing panoramic features into the original representation stream, PanoFuse maintains semantic and geometric contexts separately and routes them through DSGR to condition downstream state and action representations.

\noindent\textbf{Panoramic Representation.}
To provide global scene context, we leverage the pretrained DAP model~\cite{lin2026depth} to extract complementary semantic and geometric representations from the panoramic RGB observation $I_t^{\mathrm{pano}}$. Rather than using the predicted depth map directly, we extract semantic features $\mathbf{F}_t^{\mathrm{sem}}$ from its DINO backbone and geometric features $\mathbf{F}_t^{\mathrm{geo}}$ from its intermediate depth-decoder layer. Both representations are spatially aggregated by adaptive average pooling to an $8\times8$ grid, yielding
$\mathbf{Z}_t^{\mathrm{sem}}\in\mathbb{R}^{64\times1024}$ and
$\mathbf{Z}_t^{\mathrm{geo}}\in\mathbb{R}^{64\times256}$. We then map the two token sets into the VLM embedding space using separate learnable projections:
\begin{equation}
\tilde{\mathbf{Z}}_t^{\mathrm{sem}}
= \mathbf{Z}_t^{\mathrm{sem}}\mathbf{W}_{\mathrm{sem}}, \qquad
\tilde{\mathbf{Z}}_t^{\mathrm{geo}}
= \mathcal{N}(\mathbf{Z}_t^{\mathrm{geo}})\mathbf{W}_{\mathrm{geo}},
\label{eq:pano_projection}
\end{equation}
where $\mathbf{W}_{\mathrm{sem}}\in\mathbb{R}^{1024\times d}$ and $\mathbf{W}_{\mathrm{geo}}\in\mathbb{R}^{256\times d}$ are learnable projections, $d$ is the VLM embedding dimension, and $\mathcal{N}$ denotes per-token normalization.
The projected tokens form the semantic and geometric
context streams, respectively.

\noindent\textbf{Decoupled Semantic-Geometric Routing.}
Rather than directly mixing heterogeneous semantic and geometric representations, we organize the multimodal tokens into four functional blocks, as illustrated in Fig.~\ref{fig:pipeline}: semantic context $\mathcal{B}_0$, geometric context $\mathcal{B}_1$, state context $\mathcal{B}_2$, and action context $\mathcal{B}_3$.
We introduce Decoupled Semantic-Geometric Routing (DSGR) to control information flow among these blocks through the block-level visibility matrix
\begin{equation}
\mathbf{M} =
\begin{bmatrix}
1 & 0 & 0 & 0 \\
0 & 1 & 0 & 0 \\
1 & 1 & 1 & 0 \\
1 & 1 & 1 & 1
\end{bmatrix},
\label{eq:routing_mask}
\end{equation}
where rows correspond to query blocks and columns to key/value blocks.
Let $N$ denote the total number of input tokens and $b_i\in\{0,1,2,3\}$ the block assignment of token $i$. We expand $\mathbf{M}$ to the token-level attention mask $\mathbf{S}\in\{0,1\}^{N\times N}$ according to
\begin{equation}
\mathbf{S}
=
\left[
M_{b_i,b_j}
\right]_{i,j=1}^{N},
\label{eq:token_mask}
\end{equation}
where $S_{ij}=1$ allows query token $i$ to attend to key/value token $j$. Invalid or padded tokens are additionally masked out.

DSGR uses a fixed block-level visibility pattern, while attention weights within the permitted connections remain input-dependent. The semantic block contains wrist-image, language, and panoramic semantic tokens, whereas the geometric block contains panoramic geometric tokens. The two blocks cannot directly attend to each other, but both are independently accessible to the state and action blocks. Here, ``decoupled'' specifically refers to this interaction structure rather than an assumption that the extracted features contain exclusively semantic or geometric information.

\noindent\textbf{Training Objective.}
Following $\pi_0$~\cite{black2025pi_0}, we retain its conditional
flow-matching objective and modify only the multimodal conditioning structure through DSGR. Given a ground-truth action chunk
$\mathbf{A}_t=[\mathbf{a}_t,\ldots,\mathbf{a}_{t+H-1}]$, Gaussian noise $\boldsymbol{\epsilon}\sim\mathcal{N}(0,\mathbf{I})$, and a flow timestep $\tau$, we construct the noisy action chunk as
$\mathbf{A}_t^{\tau}=\tau\mathbf{A}_t+(1-\tau)\boldsymbol{\epsilon}$,
with target velocity $\mathbf{u}_{\tau}=\mathbf{A}_t-\boldsymbol{\epsilon}$. The multimodal representations are processed under the DSGR token-level attention mask $\mathbf{S}$ to condition the action expert. The resulting velocity predictor is optimized using
\begin{equation}
\mathcal{L}_{\mathrm{FM}}
=
\mathbb{E}_{\mathbf{A}_t,\boldsymbol{\epsilon},\tau}
\left[
\left\|
\mathbf{v}_{\theta}
\left(
\mathbf{A}_t^{\tau},
\tau,
\mathbf{o}_t,
l;
\mathbf{S}
\right)
-
\mathbf{u}_{\tau}
\right\|_2^2
\right].
\end{equation}

\section{Teleoperation System and Dataset}
\label{sec:teleop_dataset}

\begin{figure}[!t]
    \centering
    \includegraphics[width=\columnwidth]{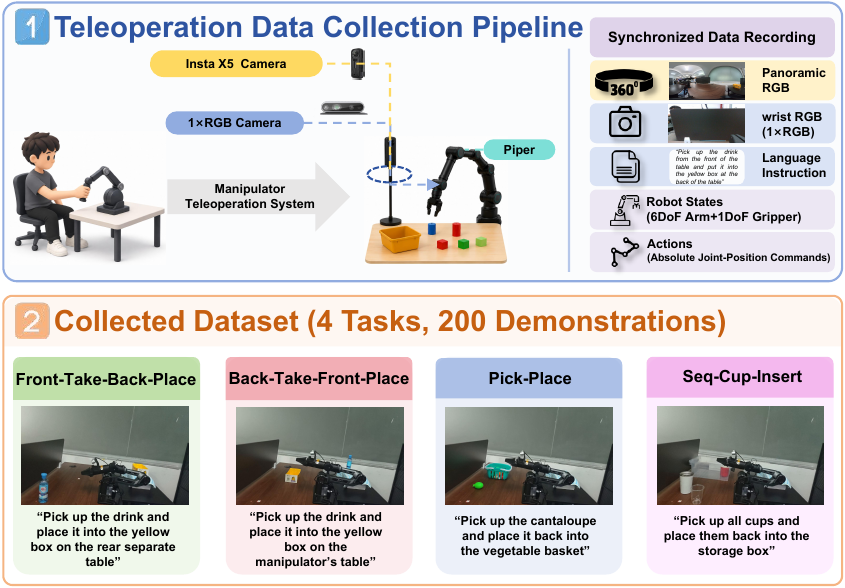}
    \vskip-1ex
    \caption{
        \textbf{Overview of our teleoperation data collection pipeline and dataset.}
        The system synchronously records panoramic RGB and wrist-view observations, language instructions, robot states, and actions during human teleoperation. We collect $200$ expert demonstrations across four real-world manipulation tasks, with $50$ trajectories per task.
    }
    \label{fig:data_collection}
    \vskip-4ex
\end{figure}

As illustrated in Fig.~\ref{fig:data_collection}, we establish a
leader--follower teleoperation system with synchronized panoramic and
wrist-view sensing for real-world manipulation data collection.

\subsection{Teleoperation System}
\label{sec:teleoperation}

\noindent\textbf{Leader--Follower Interface.}
We adopt the open-source U-ARM teleoperation framework~\cite{zou2025u}, which provides a low-cost leader--follower interface for robot manipulators. Our operator controls the robot through a seven-servo leader arm, where six servo channels correspond to the manipulator joints and an additional servo controls the gripper. At the beginning of each teleoperation episode, the
leader configuration is referenced to its initial pose. Subsequent leader-arm motion is represented relative to this reference configuration and mapped to absolute joint-position commands for the follower robot. Rather than modifying the U-ARM hardware, we integrate the leader interface with the AgileX Piper through a follower-side controller implemented using the Piper API.

\noindent\textbf{Robot Platform and Sensing.}
The follower robot is a 6-DoF AgileX Piper manipulator equipped with a 1-DoF gripper. The robot is controlled at $30$~Hz using absolute joint-position targets.
For visual perception, we employ an Insta360 X5 camera to capture a $1024\times512$ equirectangular panoramic RGB observation of the workspace. The panoramic camera is placed beside the manipulation platform to provide continuous $360^\circ$ scene coverage. 
An Intel RealSense D435i camera is mounted on the robot wrist and records $640\times480$ RGB observations, providing local visual information around the end effector and manipulated objects.

\subsection{Data Collection Protocol and Dataset}
\label{sec:dataset}

\noindent\textbf{Collection Protocol and Data Modalities.}
During each demonstration, the human operator controls the Piper through the leader arm while the recording system synchronously captures visual observations, robot states, and executed actions. Each trajectory is associated with a single natural-language task instruction. At timestep $t$, a recorded sample can be represented as
\begin{equation}
    d_t =
    \left(
    I_t^{\mathrm{pano}},
    I_t^{\mathrm{wrist}},
    l,
    s_t,
    a_t
    \right),
\end{equation}
where $I_t^{\mathrm{pano}}$ and $I_t^{\mathrm{wrist}}$ denote the panoramic and wrist-view RGB observations, respectively, $l$ is the trajectory-level language instruction, $s_t$ is the robot proprioceptive state, and $a_t$ is the executed robot action. Robot states and actions are represented by six arm joint positions together with one gripper dimension. Data are recorded at a control frequency of $30$~Hz.

\noindent\textbf{PanoFuse-Data.}
Using the teleoperation platform, we collect 200 successful expert demonstrations over four real-world manipulation tasks, with 50 trajectories per task. The resulting dataset contains more than $100$K synchronized timesteps.
The four tasks are designed to systematically cover both local manipulation and cross-workspace interaction:
\textit{Front-Take-Back-Place},
\textit{Back-Take-Front-Place},
\textit{Pick-Place}, and
\textit{Seq-Cup-Insert}.
Front-Take-Back-Place requires transferring a drink from the front workspace to a yellow box located on a separate rear table, while
Back-Take-Front-Place reverses this spatial relation by transferring the drink from the rear workspace to the manipulator table.
Pick-Place requires picking a cantaloupe and returning it to the vegetable basket. Seq-Cup-Insert requires sequentially collecting multiple cups and placing them into the storage box.
These tasks cover heterogeneous spatial dependencies, ranging from local
pick-and-place manipulation to cross-view object transfer and sequential multi-object manipulation.

\section{Experiments}
\subsection{Experimental Setup}
We evaluate four standard tasks and three generalization settings, abbreviating Front-Take-Back-Place, Back-Take-Front-Place, and Seq-Cup-Insert as F$\rightarrow$B, B$\rightarrow$F, and Cup-Insert, respectively. All generalization settings use F$\rightarrow$B: Novel-Object replaces the standard target bottle with a single unseen laboratory bottle of a different shape and size; Unseen-Background covers the robot's table with a white cloth used only at evaluation; and Distractor adds two bottles and two fruit models near the target, plus one bottle near the placement box, altering target visibility. The latter two settings retain the standard target bottle.

Each method undergoes $10$ real-world trials per setting, with the same initial object arrangement restored across trials and methods. Success requires completing the entire task without human intervention; for bottle-transfer tasks, the target bottle must be placed inside the designated box and released.

\noindent\textbf{Evaluation Metrics and Baselines.}
We use task success rate as the primary evaluation metric and report both per-setting success rates and the average success rate across all seven evaluation settings. We compare PanoFuse against four baselines derived from two pretrained $\pi_0$ variants: $\pi_0$-base~\cite{black2025pi_0} and $\pi_0$-fast~\cite{pertsch2025fastefficientactiontokenization}.

The standard $\pi_0$-base and $\pi_0$-fast baselines use two global perspective RGB cameras and one wrist-mounted RGB camera, and are trained on a separately collected set of $200$ demonstrations covering the same four tasks.
For $\pi_0$-base w/ Pano and $\pi_0$-fast w/ Pano, the panoramic RGB observation replaces the two global perspective views, while the wrist view is retained.
Both panorama-augmented baselines and PanoFuse are trained on the same $200$ recorded panoramic demonstrations, with $50$ demonstrations per task. 
The panorama-augmented baselines process the raw panoramic image through their original visual pathway, without DAP feature extraction or DSGR.
These comparisons assess the benefit of the proposed panoramic representation and routing under shared panoramic observations and training trajectories.

\begin{figure*}[!t]
    \centering
    \includegraphics[width=\textwidth]{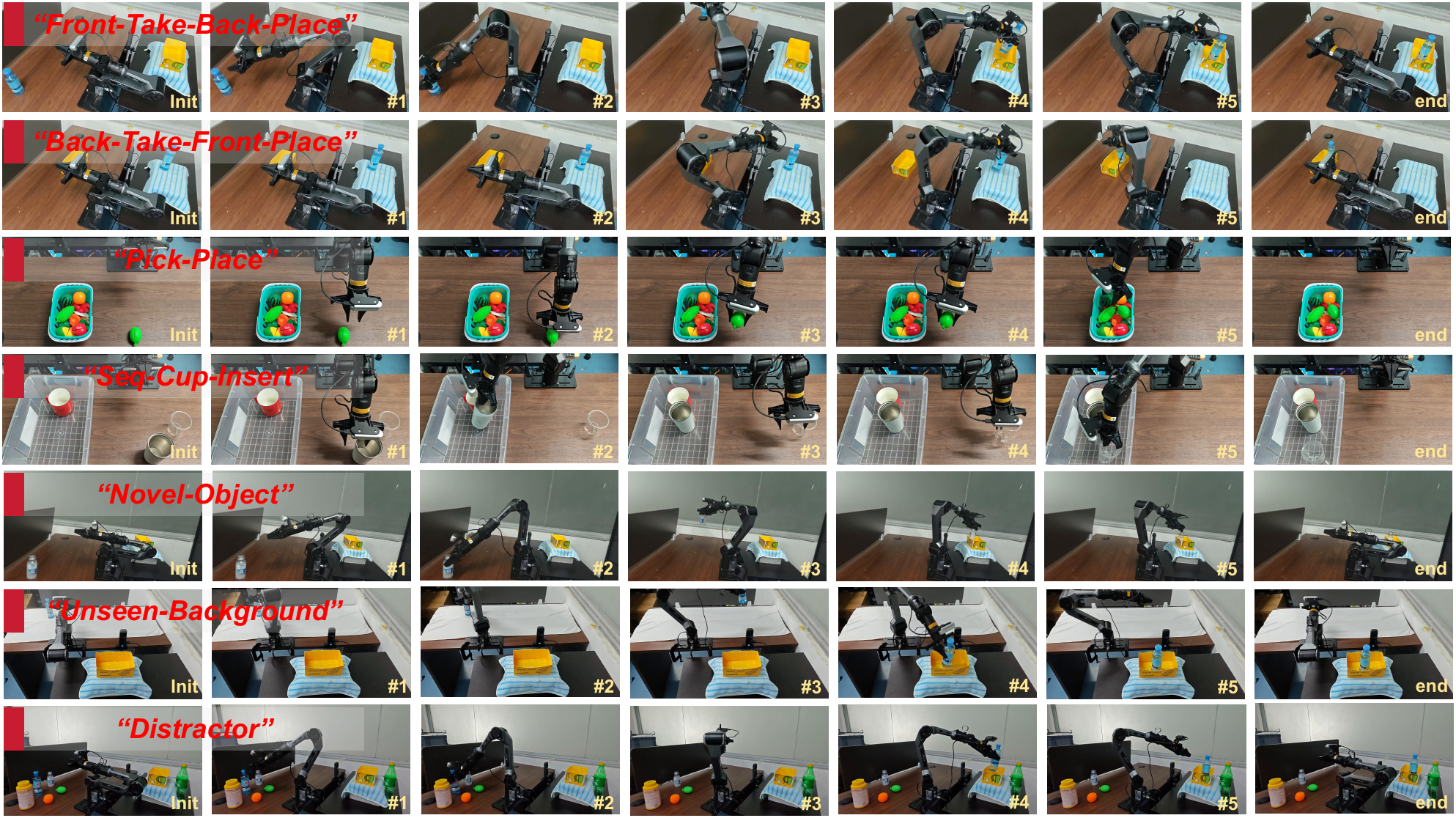}
    \vskip-1ex
    \caption{Representative real-world closed-loop rollouts of PanoFuse across seven evaluation settings. PanoFuse leverages panoramic semantic and geometric context for global scene understanding and task-relevant spatial reasoning, enabling robust manipulation under cross-workspace, novel-object, unseen-background, and distractor-rich scenarios.}
    \label{fig:tasks}
    \vskip-2ex
\end{figure*}

\begin{table*}[!t]
    \centering
    \caption{Main results across seven evaluation settings.}
    \label{tab:main_results}
    \vskip-1ex

    \small
    \setlength{\tabcolsep}{7pt}

    \begin{tabular}{c|c|c|c|c|c|c|c|c}
        \toprule

        \textbf{Method} &
        \textbf{F$\rightarrow$B} &
        \textbf{B$\rightarrow$F} &
        \textbf{Pick-Place} &
        \textbf{Cup-Insert} &
        \textbf{Novel Obj.} &
        \textbf{Unseen BG.} &
        \textbf{Distractor} &
        \textbf{Avg.} \\

        \midrule

        $\pi_0$-base~\cite{black2025pi_0}
        & 40.0\% & 50.0\% & 50.0\% & \textbf{60.0\%} & 0.0\% & 0.0\% & 0.0\% & 28.6\% \\

        $\pi_0$-base w/ Pano
        & 30.0\% & 20.0\% & 20.0\% & 20.0\% & 0.0\% & 0.0\% & 0.0\% & 12.9\% \\

        $\pi_0$-fast~\cite{pertsch2025fastefficientactiontokenization}
        & \textbf{50.0\%} & 60.0\% & 50.0\% & 50.0\% & 0.0\% & 0.0\% & 0.0\% & 30.0\% \\

        $\pi_0$-fast w/ Pano
        & 20.0\% & 20.0\% & 10.0\% & 20.0\% & 0.0\% & 0.0\% & 0.0\% & 10.0\% \\

        \midrule

        \textbf{PanoFuse (Ours)}
        & \textbf{50.0\%}
        & \textbf{70.0\%}
        & \textbf{60.0\%}
        & \textbf{60.0\%}
        & \textbf{50.0\%}
        & \textbf{40.0\%}
        & \textbf{40.0\%}
        & \textbf{52.9\%} \\

        \bottomrule
    \end{tabular}
    \vskip-4ex
\end{table*}

\noindent\textbf{Implementation Details.}
We build PanoFuse upon the pretrained $\pi_0$ VLA framework~\cite{black2025pi_0} initialized from the $\pi_0$-base checkpoint, using PaliGemma with a Gemma-2B language backbone and SigLIP for conventional RGB observations. The action expert uses a Gemma-300M backbone. 
For panoramic perception, we employ the pretrained DAP model~\cite{lin2026depth} as a frozen
feature extractor, with its semantic and geometric representations projected into the PaliGemma embedding space. 
The robot state and action each contain six arm joint positions and one gripper dimension. 
The policy predicts horizon-$50$ action chunks following the flow-matching formulation of $\pi_0$, and uses $10$ flow-integration steps
during inference.

\noindent\textbf{Training Configuration.}
These training settings are kept consistent across all compared methods where applicable. The $\pi_0$-base and $\pi_0$-fast variants are initialized from their corresponding pretrained checkpoints. The panoramic semantic and geometric representations are precomputed offline using the pretrained DAP model and remain fixed throughout policy training, reducing computational overhead and ensuring consistent panoramic representations. Before training, we compute dataset-level normalization statistics separately for the robot proprioceptive states and actions from the training demonstrations, and apply them consistently during training and inference. Each policy is trained for $20K$ optimization steps with a batch size of $4$ using bfloat16 precision. Both training and real-robot inference are performed on a single NVIDIA RTX 3090 GPU.

\subsection{Real-World Experiments}

Table~\ref{tab:main_results} summarizes the closed-loop real-robot results across four standard manipulation tasks and three generalization settings, with representative rollouts shown in Fig.~\ref{fig:tasks}. PanoFuse achieves the highest average success rate of $52.9\%$, outperforming $\pi_0$-base and $\pi_0$-fast by $24.3$ and $22.9$ percentage points, respectively. PanoFuse also maintains strong performance across all four standard tasks, indicating that panoramic context benefits manipulation across
different spatial configurations.

Notably, directly feeding raw equirectangular panoramas into the original visual pathway does not provide the same benefit. $\pi_0$-base \textit{w/} Pano and $\pi_0$-fast \textit{w/} Pano achieve only $12.9\%$ and $10.0\%$ average success, respectively, substantially below their standard counterparts. These results suggest that broader visual coverage alone is insufficient; how panoramic context is represented and integrated plays a critical role in effectively exploiting panoramic observations.

The largest performance gains occur in the three generalization settings derived from F$\rightarrow$B. PanoFuse achieves success rates of 50.0\%, 40.0\%, and 40.0\% under Novel-Object, Unseen-Background, and Distractor, respectively, whereas none of the four baselines succeeds in the 10 trials reported for each setting. These results support improved robustness under the evaluated bottle-instance, tabletop-appearance, and distractor changes.

\subsection{Ablation Studies}
All ablation variants are based on the $\pi_0$ model. The panoramic representation and attention routing ablations are evaluated on the four standard manipulation tasks, with 10 closed-loop real-robot trials per task for each variant. The panoramic FoV ablation is evaluated separately under the three generalization settings.

\begin{figure*}[!t]
    \centering
    \includegraphics[width=\textwidth]{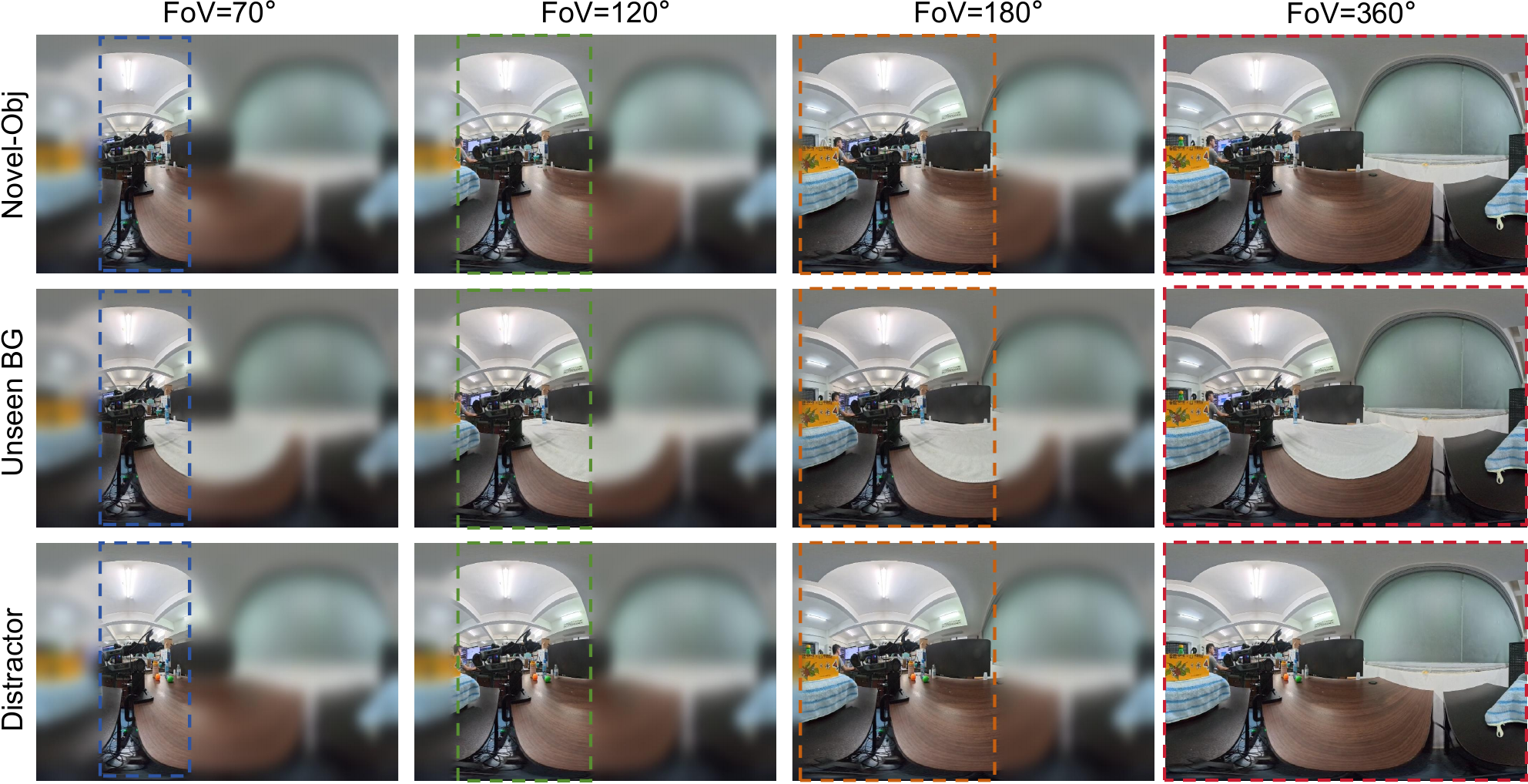}
    \vskip-1ex
    \caption{Visualization of the panoramic FoV ablation at inference time. Columns correspond to effective horizontal FoVs of $70^\circ$, $120^\circ$, $180^\circ$, and $360^\circ$. Regions outside the specified FoV are strongly blurred before panoramic feature extraction. As the FoV increases, more fine-grained scene information becomes available to PanoFuse.}
    \label{fig:FoV}
    \vskip-3ex
\end{figure*}

\noindent\textbf{Panoramic Representation Ablation.}
We investigate the contribution of panoramic semantic and geometric information by comparing the full PanoFuse model with variants that retain only the semantic or geometric panoramic representation. As shown in Table~\ref{tab:pano_modality}, using either representation alone leads to a substantial performance drop, achieving average success rates of $22.5\%$ and $20.0\%$ for the semantic-only and geometric-only variants, respectively. 
In contrast, PanoFuse achieves an average success rate of $60.0\%$, consistently outperforming both single-representation variants across all four manipulation tasks. 
These results demonstrate the complementary roles of panoramic semantic and geometric representations and highlight the importance of jointly exploiting both for effective panoramic perception.

\begin{table}[!t]
    \centering
    \caption{
        Ablation study on panoramic representations.
    }
    \label{tab:pano_modality}
    \vskip-1ex
    
    \footnotesize
    \setlength{\tabcolsep}{2.5pt}

    \begin{tabular}{c|c|c|c|c|c}
        \toprule
        \textbf{Method}
        & \textbf{F$\rightarrow$B}
        & \textbf{B$\rightarrow$F}
        & \textbf{Pick-Place}
        & \textbf{Cup-Insert}
        & \textbf{Avg.} \\
        \midrule
        Semantic Only
        & 30.0\% & 20.0\% & 20.0\% & 20.0\% & 22.5\% \\

        Geometric Only
        & 20.0\% & 20.0\% & 10.0\% & 30.0\% & 20.0\% \\

        \midrule
        \textbf{Ours}
        & \textbf{50.0\%}
        & \textbf{70.0\%}
        & \textbf{60.0\%}
        & \textbf{60.0\%}
        & \textbf{60.0\%} \\
        \bottomrule
    \end{tabular}
    \vskip-4ex
\end{table}

\noindent\textbf{Attention Routing Ablation.}
We evaluate the effectiveness of the routing structure induced by DSGR
against two alternative attention-routing configurations.
Unrestricted Attention allows all token blocks to attend to one another without structural constraints. Partial Block-wise Attention retains the same downstream routing structure but additionally allows direct interactions between the semantic and geometric blocks. In contrast, DSGR keeps semantic and geometric representations separately contextualized while allowing the state and action blocks to access both information sources.

As shown in Table~\ref{tab:attention_ablation}, Unrestricted Attention fails to complete any of the four tasks, while Partial Block-wise Attention achieves an average success rate of only $10.0\%$. Our structured block-wise attention substantially improves the average success rate to $60.0\%$, with consistent improvements across all four tasks. 
These results suggest that simply enabling direct interactions among heterogeneous representations is insufficient. Instead, explicitly separating semantic and geometric contextualization while exposing both to downstream state and action representations leads to more effective multimodal information integration for action generation.

\noindent\textbf{Panoramic FoV Ablation.}
We further investigate the contribution of global scene coverage by systematically varying the horizontal FoV of the panoramic observation at inference time to more comprehensively assess its influence. We evaluate $70^\circ$, $120^\circ$, $180^\circ$, and $360^\circ$ FoVs under the three generalization settings, with the FoV centered at the viewing direction and all other inference settings unchanged to isolate the effect of coverage for a controlled comparison. As illustrated in Fig.~\ref{fig:FoV}, regions outside the specified FoV are strongly blurred with smooth boundary transitions before DAP feature extraction, thereby progressively varying the amount of panoramic context available to the policy.

\begin{table}[t]
    \centering
    \caption{
        Ablation study on attention routing.
    }
    \label{tab:attention_ablation}
    \vskip-1.0ex
    
    \footnotesize
    \setlength{\tabcolsep}{2.5pt}

    \begin{tabular}{c|c|c|c|c|c}
        \toprule
        \textbf{Method}
        & \textbf{F$\rightarrow$B}
        & \textbf{B$\rightarrow$F}
        & \textbf{Pick-Place}
        & \textbf{Cup-Insert}
        & \textbf{Avg.} \\
        \midrule
        Unrestricted
        & 0.0\% & 0.0\% & 0.0\% & 0.0\% & 0.0\% \\

        Partial Block-wise
        & 10.0\% & 10.0\% & 20.0\% & 0.0\% & 10.0\% \\

        \midrule
        \textbf{Ours}
        & \textbf{50.0\%}
        & \textbf{70.0\%}
        & \textbf{60.0\%}
        & \textbf{60.0\%}
        & \textbf{60.0\%} \\
        \bottomrule
    \end{tabular}
    \vskip-1ex
\end{table}

As shown in Table~\ref{tab:FoV_ablation}, generalization performance improves substantially as broader scene context becomes available. The average success rate increases from $6.7\%$ at $70^\circ$ to $10.0\%$ at $120^\circ$ and $26.7\%$ at $180^\circ$, while the full $360^\circ$ PanoFuse achieves $43.3\%$.
The improvement is particularly pronounced under Novel-Object, where the success rate increases from $0\%$ at $70^\circ$ and $120^\circ$ to $10.0\%$ at $180^\circ$ and $50.0\%$ with the full panorama. 
These results indicate that the generalization capability of PanoFuse benefits from access to broader scene-level context, supporting the use of full panoramic observations rather than restricted local views.

\begin{table}[!t]
    \centering
    \caption{Ablation study on panoramic FoV at inference time.}
    \label{tab:FoV_ablation}
    \vskip-1.0ex

    \footnotesize
    \setlength{\tabcolsep}{2.5pt}

    \begin{tabular}{c|c|c|c|c}
        \toprule
        \textbf{Method}
        & \textbf{Novel Obj.}
        & \textbf{Unseen BG.}
        & \textbf{Distractor}
        & \textbf{Avg.} \\
        \midrule

        FoV=$70^\circ$
        & 0.0\%
        & 10.0\%
        & 10.0\%
        & 6.7\% \\

        FoV=$120^\circ$
        & 0.0\%
        & 20.0\%
        & 10.0\%
        & 10.0\% \\

        FoV=$180^\circ$
        & 10.0\%
        & \textbf{40.0\%}
        & 30.0\%
        & 26.7\% \\

        \midrule

        \textbf{Ours ($360^\circ$)}
        & \textbf{50.0\%}
        & \textbf{40.0\%}
        & \textbf{40.0\%}
        & \textbf{43.3\%} \\

        \bottomrule
    \end{tabular}
    \vskip-4ex
\end{table}

\section{Conclusion}
In this work, we presented PanoFuse, a panorama-enhanced vision-language-action framework for real-world robotic manipulation.
PanoFuse extracts complementary semantic and geometric representations from panoramic observations and integrates them with local visual, language, proprioceptive, and action representations through structured block-wise attention.
Real-world experiments across seven evaluation settings demonstrate that PanoFuse achieves the highest overall average success rate, while exhibiting stronger generalization under unseen visual conditions.
Ablation studies further confirm the complementary roles of panoramic semantic and geometric representations and the importance of structured information integration. 
These results show that effective panoramic perception requires not only broader visual coverage, but also representing and integrating global context in a form suitable for robust action generation. Future work could extend PanoFuse to heterogeneous robot embodiments and explore richer multimodal sensing to further improve cross-platform generalization and robustness in diverse manipulation environments.

\bibliographystyle{IEEEtran}
\bibliography{bib}

\end{document}